\pdfoutput=1
\documentclass[11pt]{article}
\usepackage{acl}

\usepackage{times}
\usepackage{latexsym}
\usepackage{multirow}
\usepackage{booktabs}
\usepackage{soul}
\usepackage{hyperref}
\usepackage{enumitem} 
\usepackage[inkscapeformat=png]{svg}
\usepackage{siunitx} 
\usepackage{graphicx}
\usepackage{amsmath} 
\usepackage{caption} 
\usepackage{algpseudocode} 
\usepackage{algorithm} 
\usepackage[T1]{fontenc}
\usepackage{todonotes}

\usepackage[utf8]{inputenc}
\usepackage{microtype}
\usepackage{graphicx}
\usepackage{amsmath}
\usepackage{amssymb}
\usepackage{arydshln}
\usepackage{tabularx}
\usepackage[utf8]{inputenc}

\definecolor{ForestGreen}{RGB}{34,139,34}

\usepackage{array} 
\usepackage{colortbl} 
\usepackage{pdflscape} 

\newcolumntype{Y}{>{\centering\arraybackslash}X}

\usepackage{booktabs}
\usepackage{todonotes}
\usepackage{multirow}
\usepackage{multicol}
\usepackage{graphicx}
\usepackage{subcaption}
\usepackage{colortbl}

\title{\textbf{\texttt{CSEval}}: Towards Automated, Multi-Dimensional, and Reference-Free Counterspeech Evaluation using Auto-Calibrated LLMs}

\author{Amey Hengle$^{1}$\thanks{* Equal contribution}, Aswini Kumar$^{1*}$, \textbf{Anil Bandhakavi}$^2$, \textbf{Tanmoy Chakraborty}$^1$\\
$^1${Indian Institute of Technology Delhi, India};
$^2${Logically.ai} \\
\small {
\{
\texttt{ameyhengle22},
\texttt{aswinikumarpadhi1995}
\}
\texttt{@gmail.com}
}\\
\small
{\tt tanchak@iitd.ac.in}
}

\begin{document}
\maketitle

\begin{abstract}
Counterspeech has emerged as a popular and effective strategy for combating online hate speech, sparking growing research interest in automating its generation using language models. However, the field still lacks standardised evaluation protocols and reliable automated evaluation metrics that align with human judgement. Current automatic evaluation methods, primarily based on similarity metrics, do not effectively capture the complex and independent attributes of counterspeech quality, such as contextual relevance, aggressiveness, or argumentative coherence. This has led to an increased dependency on labor-intensive human evaluations to assess automated counter-speech generation methods. To address these challenges, we introduce \texttt{CSEval}, a novel dataset and framework for evaluating counterspeech quality across four dimensions: \textit{contextual-relevance}, \textit{aggressiveness}, \textit{argument-coherence}, and \textit{suitableness}. Furthermore, we propose {\em Auto-Calibrated COT for Counterspeech Evaluation} (\texttt{Auto-CSEval}), a prompt-based method with auto-calibrated chain-of-thoughts (CoT) for scoring counterspeech using large language models. Our experiments show that \texttt{Auto-CSEval} outperforms traditional metrics like ROUGE, METEOR, and BertScore in correlating with human judgement, indicating a significant improvement in automated counterspeech evaluation. \footnote{\textit{\color{red} Warning: The content in this paper may be upsetting or offensive.}}
\end{abstract}
\begin{figure}[t]
\includegraphics[width=\columnwidth]{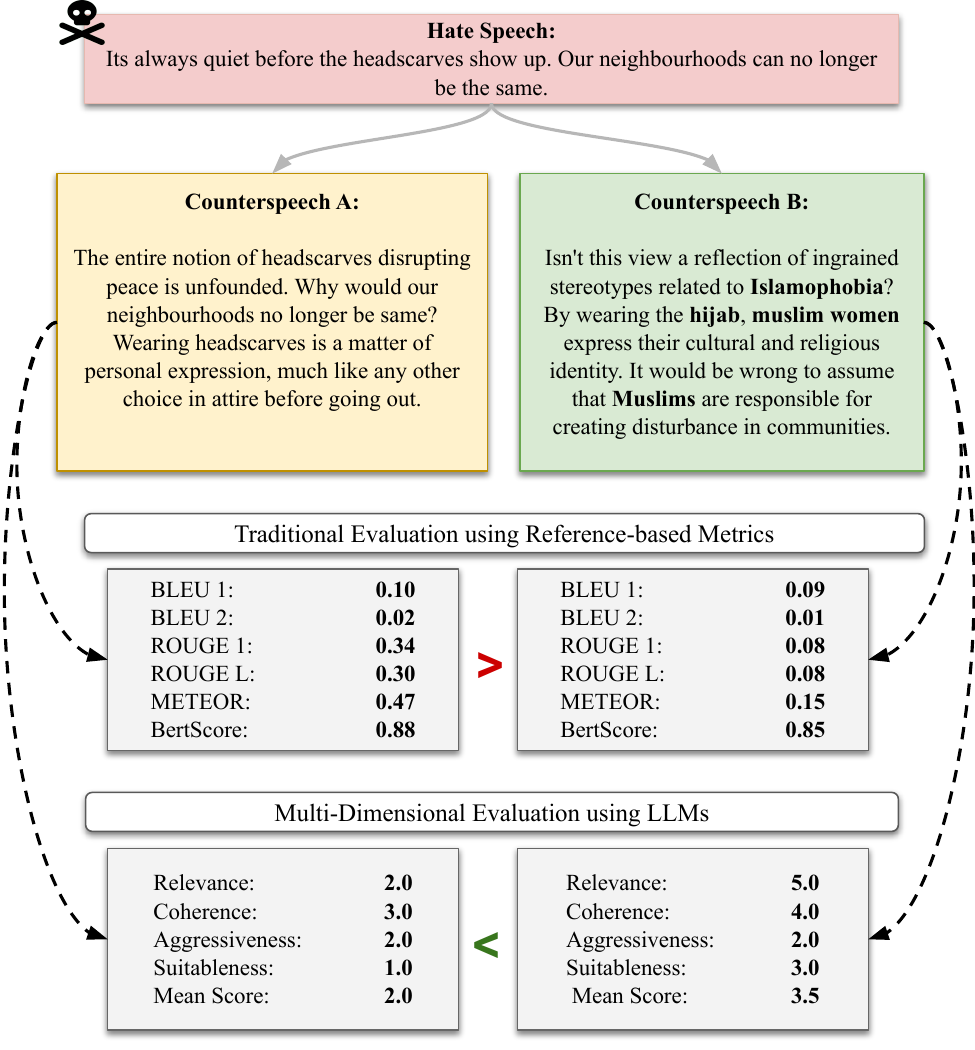}
\caption{An example comparing classical evaluation (ROUGE, METEOR, etc.) vs LLM-based multidimensional evaluation for two counterspeech, A and B. We observe that while B is more relevant and addresses the implied bias expressed in the hate speech, it is scored lower than A by traditional metrics. In contrast, LLM-based multidimensional evaluation aligns more with human judgement, scoring B higher than A.}
\label{fig:introduction_figure}
\vspace{-5mm}
\end{figure}
\section{Introduction}
Online hate speech ({HS}) is on the rise in social media, making it a hostile platform for targeted individuals. Counterspeech (CS) provides an efficient way to combat hate speech with the help of constructive statements \cite{benesch2016-considerations-for-succesful, chandrasekharan2017you} without violating the speaker's freedom of speech \cite{schieb2016governing, wright2017vectors}.  \textcolor{black}{Moreover, counterspeech has popularly emerged as an effective strategy for countering hateful comments without content moderation and deletion \cite{leetaru-2019, saha2019complex}.} Due to the tediousness of human-generated counterspeech, the natural language generation ({NLG}) approach motivates the researchers to focus on generating automated counterspeech \cite{mathew2019thou, qian2019benchmark, chung-et-al-2023-understanding-counterspeech, fanton2021human, bonaldi-etal-2022-human, hengle2024intentconditioned}.

The rapid development of automated counterspeech generation methods using pre-trained language models (PLMs) calls for a high-quality evaluation of generated counterspeech. However, the evaluation process is still dominated by traditional similarity-based NLG metrics like BLEU \cite{papineni2002bleu}, ROUGE \cite{rouge-score-lin-2004}, and METEOR \cite{meteor-score-banerjee-lavie-2005} These metrics focus on surface-level text differences and often fall short in semantic aspects \cite{freitag-etal-2020-bleu}. Additionally, other methods based on neural embeddings \cite{bert-score-zhang2020, yuan2021BartScore}  are inflexible and limited in scope \cite{freitag-etal-2021-experts}. Moreover, these metrics show low alignment with human judgment, especially for open-ended generation tasks, where there is a possibility of multiple correct outputs \cite{guan-huang-2020-union, zhong-etal-2022-UniEVAL, liu2023-GEVAL}. This problem is particularly pertinent in counterspeech evaluation, where the reference text (\texttt{HS}) is characterized by short, indirect, and implied expressions of hate \cite{hengle2024intentconditioned}. 

We support our argument by providing an example in Figure \ref{fig:introduction_figure}, where for a given hate speech, we contrast two counterspeech, A and B, generated by GPT3.5 \cite{ouyang2022-instuctGPT}. We observe while A makes valid arguments, B is more effective and relevant in countering the hate speech compared to A, as it accurately refutes the underlying {\textit{Islamophobic sentiment}} and addresses both the stereotype and broader context of cultural and religious identity. However, all traditional methods consistently score A higher than B, given their lexical and semantic proximity to the reference text. These shortcomings motivate the need for more nuanced, multi-dimensional, and human-aligned evaluation methods for automated counterspeech generation.

The emergent capabilities of LLMs such as instruction tuning \cite{wei2022finetuned-models-are-zero-shot}, and preference tuning through human feedback \cite{ouyang2022-instuctGPT}, and as Chain-of-Thought (CoT) reasoning \cite{wei2023chainofthought} have presented a promising direction towards NLG evaluation based on LLMs. Recent studies have proposed using LLMs as reference-free NLG evaluators directly, offering a better human-aligned evaluation compared to traditional methods. \cite{wang2023chatgpt_a_good_NLG_evaluator, fu2023GPTScore, liu2023-GEVAL, jones2024-multiaspectframeworkcounternarrative}. However, the validity and reliability of using LLMs as NLG evaluators have yet to be systematically investigated in automated counterspeech generation \cite{chung-et-al-2023-understanding-counterspeech}. Specifically, there is an issue with how automated metrics are evaluated themselves. \citet{jones2024-multiaspectframeworkcounternarrative} took an initial step in this direction by proposing a dataset and framework for reference-free counterspeech evaluation. However, they lack expert annotations and have a relatively small test size (180 samples) \cite{jones2024-multiaspectframeworkcounternarrative}. Thus, there is no dataset with expert human judgements evaluating specific quality aspects of model-generated counterspeech. 

We address these gaps in complementary ways. First, we build \textbf{\texttt{CSEval}}, a large and diverse collection (in terms of model-types) of human judgements of model-generated counterspeech across four quality aspects: \textit{contextual-relevance}, \textit{aggressiveness}, \textit{argument-coherence}, and \textit{suitableness}. We release aligned model outputs for five of the current state-of-the-art (SoTA) models on counterspeech generation. We benchmark \texttt{CSEval} on multiple reference-based as well as reference-free evaluation methods. Finally, we propose \textbf{Auto Calibrated Chain-of-Thoughts for Counterspeech Evaluation} (\texttt{Auto-CSEval}), a prompt-based method with auto-calibrated evaluation steps (CoT) to score each quality dimension of the counterspeech. Experimental results show that our proposed approach outperforms the traditional and LLM-based evaluation methods in terms of correlation with human judgement.

To summarize our contribution, in this paper, we -- (i) introduce \texttt{CSEval}, the largest and most diverse collection of human judgments of model-generated counterspeech across four quality aspects: context relevance, aggressiveness, argument coherence, and suitableness; (ii) benchmark the \texttt{CSEval} dataset across multiple popular automatic evaluation methods and find that none of them reliably measure the quality of counterspeech, showing poor correlation with human-judgments; (iii) propose \texttt{Auto-CSEval}, a prompt-based method with auto-calibrated CoTs to score each quality dimension of the counterspeech, and show that it outperforms the traditional and LLM-based evaluation strategies which are correlated with the human judgment \footnote{\textcolor{black}{The source code and datasets are available at https://github.com/AmeyHengle/cs-eval}}.
\section{Related Work}
There has been an increased interest in research focusing on building datasets and related resources \cite{chung2019conan, liu-etal-2019-multi, gupta-etal-2023-counterspeeches} along with automated methods for counterspeech generation \cite{zhu-bhat-2021-generate, zhang2020dialogpt, gupta-etal-2023-counterspeeches, hengle2024intentconditioned}.  Within this context, the IntentCONAN dataset \cite{gupta-etal-2023-counterspeeches} is particularly notable for reflecting the style-guided of the counterspeech generation task, providing multiple ground-truth counterspeech for a given hate speech statement. Further, some recent efforts \cite{mun2023-beyond-denouncing-hate, saha2024zeroshot} have started leveraging prompting-based strategies for counterspeech generation.

With the recent progress in this domain, automated evaluation of model-generated counterspeech remains an open problem. Generally speaking, almost all of the previous studies \cite{zhu-bhat-2021-generate, gupta-etal-2023-counterspeeches, controlgedi, hengle2024intentconditioned} extensively employ lexical overlap \cite{papineni-etal-2002-bleuScore, rouge-score-lin-2004, meteor-score-banerjee-lavie-2005}, semantic similarity \cite{bert-score-zhang2020}, and diversity \cite{li2016diversitypromoting} metrics for counterspeech evaluation. However, reference-based metrics have been consistently shown to have poor correlation with human judgements, especially for evaluating the quality of open-ended generation tasks \cite{zhong-etal-2022-UniEVAL, fu2023GPTScore, li2024leveraging}. Furthermore, these metrics are incapable of representing the aspect-level quality of a counterspeech, such as soundness, relevance, or non-toxicity \cite{chung-et-al-2023-understanding-counterspeech}. Due to these reasons, research in counterspeech generation is heavily dependent on human evaluation, which is often very costly to conduct and difficult to scale. 

With their growing reasoning and generation capabilities, LLMs are now being considered as an alternative to human evaluation, showing a high degree of correlation with human judgement on evaluating tasks like summarisation, dialog-generation, and code generation \cite{wang2023chatgpt_a_good_NLG_evaluator, liu2023-calibrating-LLM-Evaluator, fu2023GPTScore, liu2023-GEVAL, li2024leveraging}. \citet{fu2023GPTScore} propose GPTScore, an automated framework that evaluates text quality with pre-trained generative models like GPT3. \citet{kocmi2023large} propose to use GPT-based models for scoring machine translation outputs.  \citet{li2024leveraging} conduct an extensive survey on the usage of LLMs for NLG evaluation. Recently, \citet{jones2024-multiaspectframeworkcounternarrative} introduced a reference-free evaluation framework that leverages LLMs to assess counterspeech quality across multiple dimensions. They showed that LLM-based evaluation aligns more closely with human judgement compared to traditional methods. \citet{liu2023-GEVAL} propose a prompt-based method to further improve correlation with human judgement by generating detailed evaluation steps using the CoT reasoning capability of GPT-4. Our proposed approach, Auto-CSEval, builds on both these works. However, instead of relying solely on model-generated CoT reasoning like \cite{liu2023-GEVAL}, we actively calibrate the LLM using a small set of human judgements. Our method ensures a more robust and human-aligned scoring of counterspeech using LLMs. 
\section{Dataset}
We build and release \textbf{\texttt{CSEval},} a benchmark dataset for reference-free and multi-dimensional counterspeech evaluation. \texttt{CSEval} contains expert human assessments of $7,926$ model-generated \texttt{CS}, across four quality dimensions (or aspects). In this section, we give a detailed overview of the data curation process, evaluation dimensions, and the models used in \texttt{CSEval}. 

\subsection{Data Curation}
We used the publicly available IntentCONAN \cite{gupta-etal-2023-counterspeeches} dataset, which offers a diverse set of \texttt{CS} spanning multiple categories like empathy, counter-questions, fact-checking, and denouncing. We selected IntentCONAN over other counterspeech datasets \cite{chung2019conan, fanton2021humanintheloop}, as it accurately reflects the open-ended nature of counterspeech generation, accommodating multiple valid \texttt{CS} for a given \texttt{HS}. We began by randomly sampling approximately $2000$ unique \texttt{HS} instances from IntentCONAN. For each \texttt{HS} instance, we then generated a \texttt{CS} using five popular counterspeech generation models, as detailed in Appendix \ref{appendix:model_description}. We ended up creating the base corpus of \texttt{CSEval} with $2,223$ unique \texttt{HS}, $4,318$ ground-truth (reference) \texttt{CS}, and $7,926$ model-generated \texttt{CS} instances.

\subsection{Evaluation Dimensions}
We define the evaluation process across four dimensions (or aspects) of counterspeech quality $-$ relevance, aggressiveness, coherence, and suitableness. We select these dimensions as they are most frequently reported in human evaluation studies in counterspeech literature. We discuss this in detail in Appendix \ref{sec:appendix_eval_dims}.

(i) {\bf Relevance} assesses whether the counterspeech is in line with the hate speech's central theme, subject, or topic. Contextual relevance is an important counterspeech quality, especially considering the implied nature of hate speech that can confuse language models.

(ii) {\bf Coherence} measures whether a counterspeech provides specific and coherent arguments to effectively refute or counter any bias, stereotype, or prejudice expressed in the hate speech. A high score indicates that the counterspeech provides arguments that are consistent, evidence-based, and follow a clear logical flow and use. 

(iii) {\bf Aggressiveness} evaluates the level of confrontational or inflammatory content in the counterspeech, including the use of any abusive language, the intensity of disagreement, the tone, and whether it contains personal attacks. A lower score indicates less aggressive and hence more effective counterspeech. 

(iv) {\bf Suitableness} measures whether a counterspeech can be directly used without editing in a real setting. It considers a counterspeech's overall stance and potential impact on the listener.

\subsection{Models} 
\label{sec:model_descripton}
We include counterspeeches generated using both supervised and prompt-based methods. In terms of supervised methods, we include three popular counterspeech generation models -- \textbf{QUARC} \cite{gupta-etal-2023-counterspeeches},  \cite{zhu-bhat-2021-generate}, \textbf{DialoGPT} \cite{zhang2020dialogpt}, and \textbf{Generate-Prune-Select (GPS)}. As our prompting baselines, we include counterspeeches generated by two popular LLMs --\textbf{GPT-3.5-Turbo} (ChatGPT) and \textbf{GPT-4} \cite{ouyang2022-instuctGPT}. For each of them,we include both zero- and few-shot prompting baselines. Further details surrounding model training and prompting are provided under Appendix \ref{appendix:model_description}. 

\subsection{Annotation Process} 

\textcolor{black}{We recruited five expert annotators who have either published papers or completed senior theses in the domain of hate speech and counterspeech \footnote{\textcolor{black}{All annotators were aged between 20-30 years, with a gender distribution of 80\% male and 20\% female}}. We used expert annotators for our evaluation process due to the task's sensitivity and the quality issues of crowd-sourced annotations reported in previous work \cite{gillick-liu-2010-nonExpertAnnotationsRisky}.} Given a \texttt{HS}, reference \texttt{CS}, and model-generated \texttt{CS}, annotators are asked to rate the model output across each of the four dimensions on a Likert scale. Relevance, coherence, and aggressiveness are rated on a scale of $1$ to $5$, while suitableness is rated on a scale of $1$ to $3$. Higher scores indicate better quality, except for aggressiveness, where lower scores indicate better quality. Further details about the background and discussion process with the annotators can be found in Appendix \ref{sec:appendix-dataset}.

\paragraph{Inter-Annotator Agreement.} We used Krippendorff’s alpha coefficient \cite{Krippendorff2011ComputingKA} to measure the inter-annotator agreement of the expert annotations. For the first round of annotations, we obtained a decent inter-annotator interval of Krippendorff's (averaged across the four dimensions) of $0.473$. However, the 2nd round increases the overall inter-annotator agreement with an average Krippendoff coefficient of $0.681$. We also calculate the standard deviation of annotator scores within the respective groups. We plot the histogram of these statistics in Appendix Figure \ref{fig:histogram}. Here, we observe that \textit{suitableness} remains the most contentious dimension among experts. Further details about the annotation process are provided under Appendix Section \ref{sec:appendix-dataset}.

\begin{figure*}[t]
\includegraphics[width=\textwidth]{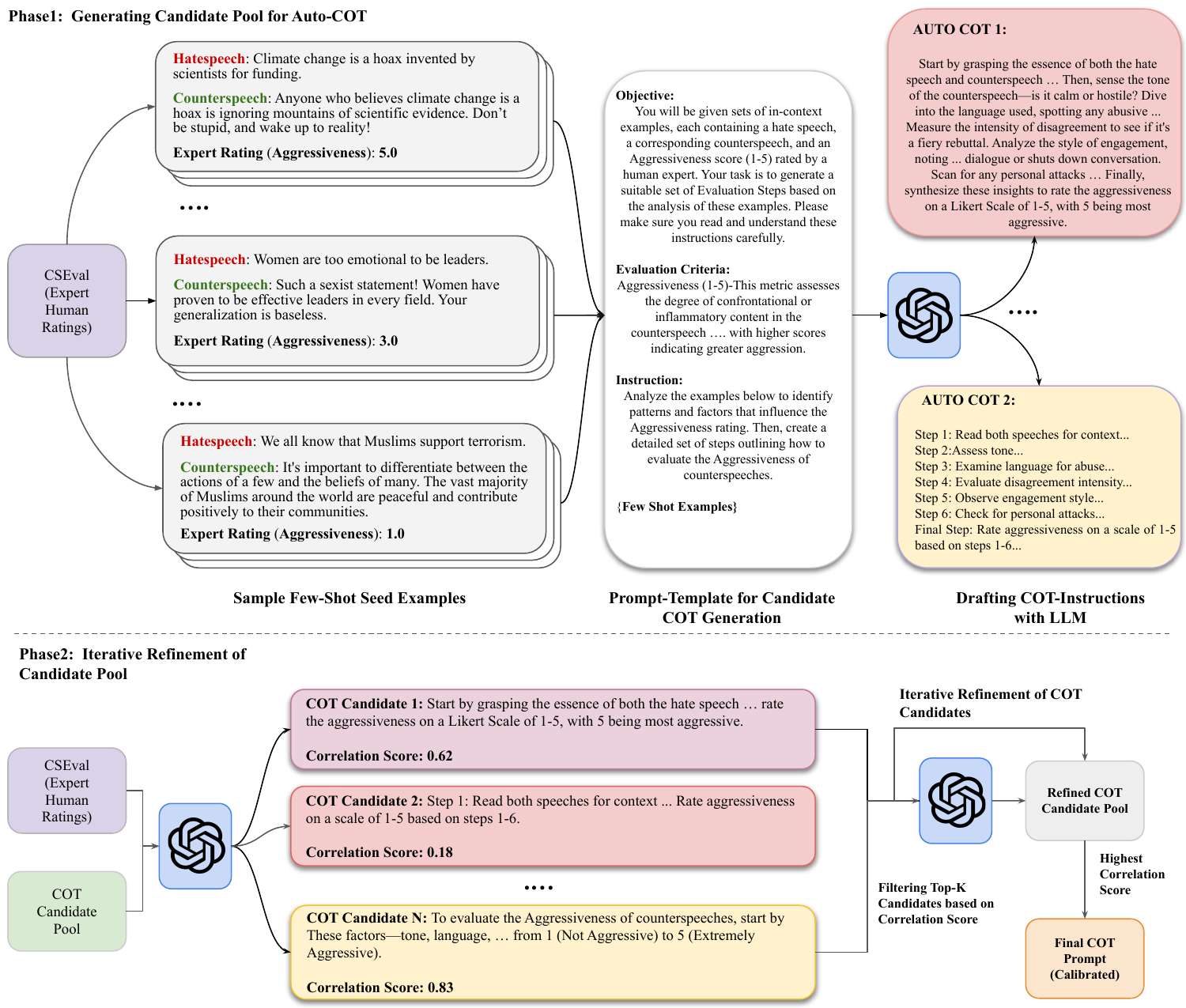}
\caption{An overview of the multi-phase auto-calibration framework of \texttt{Auto-CSEval}, including the generation and refinement of CoT instructions, evaluation criteria formulation, and the iterative calibration process aligned with expert human judgement.}
\label{fig:figure2}
\vspace{-5mm}
\end{figure*}
\section{Proposed Method}
\label{sec:proposed_methodology}
Figure \ref{fig:figure2} illustrates the overall framework of our proposed method, \textbf{\texttt{Auto-CSEval}} (\textit{Auto-Calibrated Chain-of-Thoughts for Counterspeech Evaluation}). Auto-CSEval is a prompt-based evaluator with three main components: (i) An instruction describing the evaluation task at hand, (ii) Evaluation CoT, i.e., a brief description of the evaluation aspect, and (iii) CoT that is the set of intermediate instructions describing the detailed evaluation steps for the desired evaluation CoT. To calibrate the LLM, we focus on optimising the CoT steps $T$, while keeping the instruction and evaluation CoT parts of the prompt consistent. Specifically, we mine and tune the CoT steps by constructing a small validation set, containing human-labelled $\langle$HS, CS$\rangle$ pairs. Following this, we adopt a step-by-step procedure to iteratively refine the candidate CoT. This includes two phases, i) drafting and ii) revisiting as shown in \ref{fig:figure2}. The first set of candidate CoT drafts is obtained by running inference with in-context labels using an induction prompt. These are then evaluated and filtered on expert labels and then refined to accommodate erroneous evaluations. 

\paragraph{Instruction:} The prompt is a natural language instruction that defines the evaluation task at a high level. In our study, we use a common instruction prompt for all four evaluation dimensions.
\begin{quote}
\textit{You will be given one counterspeech (also called counterspeech or counter-narrative) written for a hateful statement (hate speech). Your task is to rate the counterspeech on one metric. Please make sure you read and understand these instructions carefully.}
\end{quote}

\noindent
\textbf{Definition:}
The definition provides a brief description of the evaluation dimension. Below is an example of the definition prompt for aggressiveness. 
\begin{quote}
\textit{Evaluation Criteria:}
\textit{Aggressiveness (1-5) -- This metric assesses the degree of confrontational or inflammatory content in the counterspeech. It considers factors such as abusive language, intensity of disagreement, tone, engagement style, and the presence of personal attacks, with higher scores indicating greater aggression.}
\end{quote}

\noindent
\textbf{Auto Evaluation Steps (CoT):}
CoT is a prompt engineering technique that encourages LLMs to break down a complex task into a sequence of intermediate steps and has been shown to improve LLMs in their reasoning abilities and generate more accurate and informative responses for multi-step problems, such as arithmetic, common-sense, and symbolic reasoning \cite{wang2023chatgpt_a_good_NLG_evaluator, zheng2023judging_LLMs}. For evaluation tasks, some candidate CoTs need more detailed instructions (evaluation steps). It is generally time-consuming to  design such evaluation steps for each task manually. \citet{liu2023-GEVAL} found that an LLM can generate such evaluation steps by itself and proposed an auto-CoT approach to evaluate NLG tasks. However, this approach may result in sub-optimal and misaligned CoT where the scoring guidelines are absent and only output categorical ranges (e.g., 0-5) are provided. LLM-based evaluators are shown to suffer from insufficient prompting, resulting in inconsistent and misaligned evaluations \cite{lu2023error_chatGPT}. However, manually defining CoT instructions or evaluation steps is both challenging and time-consuming and may require an expert who has the requisite domain knowledge to establish the evaluation criteria. Furthermore, manually defining an evaluation criteria may also lead to personal bias of the evaluator. Recently, \citet{liu2023-calibrating-LLM-Evaluator} proposed a framework to auto-calibrate LLM prompts based on a validation set of human judgments. Drawing inspiration from this, we propose a data-driven approach to calibrating the most optimal CoT instructions (or evaluation steps) for a given dimension. As illustrated in Figure \ref{fig:figure2}, our method follows a two-phase procedure.

In the first phase, we begin by constructing a validation set $D^*$ of $500$ model outputs by randomly sampling from the \texttt{CSEval} dev set. This is denoted by $D$. After this step, we prompt an LLM to independently generate the scoring CoT $C$ from few-shot \texttt{HS}-\texttt{CS} exemplars. While doing this, it is also important to make sure that the generated CoT are non-repetitive. Monte-Carlo \cite{monte-carlo} trials are shown to help in countering any kind of label- or position-related bias during sampling. Therefore, following \cite{liu2023-calibrating-LLM-Evaluator}, we conduct Monte-Carlo trials and sample fewshot exemplars from the development set $D^*$, which are then used directly in the prompt. 

Given an arbitrary sample \( d_i \in D \), the prompt template for chain of thought (CoT) drafting \( T_D \), evaluation dimension \( a \) (e.g., relevance, aggressiveness), and a counterspeech-model \( LLM(\cdot) \), the quality dimension \( d_i \) is evaluated as \( \hat{s}_{i,a} = LLM(T_D(d_i, C, a)) \). With this setup in place, a corresponding candidate CoT can be inferred as follows:
\begin{equation}
\hat{C} = \underset{C}{\mathrm{argmax}} \; P_\theta (C|T_D(D_s, a))
\end{equation}

In the given equation, \( D_s = \bigcup_{i}(d_i^*, s_{i,a}) \subset D^* \) represents the subset of few-shot exemplars. On the other hand, \( P_\theta \) denotes the probability distribution modeled by the LLM's parameters \( \theta \). To allow for diversity in candidate CoT prompts, we apply temperature sampling to draw scoring CoT from the LLM, similar to \cite{liu2023-calibrating-LLM-Evaluator}. \textcolor{black}{The temperature values for temperature sampling are dynamically selected from a range of ($0$ to $0.5$), for generation diversity while curating the candidate CoT.} The prompt templates used in this process are provided in Appendix \ref{appendix:prompt_templates}. After obtaining the initial set of candidate CoTs in the first phase, we move to evaluation and refinement in the second phase.

In the second phase, we iteratively refine the candidate CoT pool obtained from the previous phase using $D^*$. The CoTs obtained in the first phase may be sub-optimal. In order to filter out high-quality candidates, we first evaluate them using the dev set $D^*$. We then sample only the best-performing candidate CoT $C$ based on their Spearman correlation scores (the higher the better). To address the low correlation between human ratings and the initial CoT, we prompt the LLM to refine its previously generated CoT through self-editing. In this process, we use examples with the lowest correlation scores (strong disagreement). \textcolor{black}{Specifically, we identify the top-3 instances with the highest absolute difference between the predicted and human scores and use these as fewshot examples for the CoT-refinement step.}. This self-editing step ensures that all candidate CoT \(C\) are progressively aligned with human ratings at each iteration. This improves diversity and helps to reduce any biases in the final prompt \cite{liu2023-calibrating-LLM-Evaluator}.
\begin{enumerate}[noitemsep,topsep=0pt]
    \item Modify: Alter parts of the generated chain-of-thought evaluation steps (CoT) with the aim of increasing its correlation with human scores.
    \item Paraphrase: If a candidate chain-of-thought (CoT) scores below the correlation threshold, paraphrase it to make it clearer and more concise.
    \item Addition of new rules or evaluation steps: If the model $LLM(T_D(d_i, C, a))$ finds new scoring rules that are not present in the current candidate CoT $C$, append $C$ with the new rules.
    \item Calibrate: Any other modifications that the LLM infers to improve correlation with human judgments.
\end{enumerate}
Figure \ref{fig:figure2} provides an overview of our two-phased calibration process. As shown in the figure, after obtaining a new candidate CoT $C$, we sample them using the validation set $D^*$. After this step, we combine the new CoT $\hat{C}$ with the pre-filtered draft CoTs. This gives us a final calibrated set of scoring rules (evaluation steps). Algorithm \ref{tab:pseudo_code} details the pseudo-code followed for CoT selection and iterative refinement.

\begin{table*}[!ht]
\begin{center}
\captionsetup{justification=centering}
\resizebox{\textwidth}{!}{
\begin{tabular}{lrrrrrrrrrr}
\toprule
\multicolumn{1}{c}{\textbf{Metrics}} & \multicolumn{2}{c}{\textbf{Context Relevance}} & \multicolumn{2}{l}{\textbf{Aggressiveness}} & \multicolumn{2}{c}{\textbf{Argument Coherence}} & \multicolumn{2}{c}{\textbf{Suitableness}} & \multicolumn{2}{c}{\textbf{AVG}} \\ 
\cmidrule(lr){2-3} \cmidrule(lr){4-5} \cmidrule(lr){6-7} \cmidrule(lr){8-9} \cmidrule(l){10-11} & {$\rho$} & {$\tau$} & {$\rho$} & {$\tau$} & {$\rho$} & {$\tau$} & {$\rho$} & {$\tau$} & {$\rho$} & {$\tau$} \\ 
\midrule
\textbf{Similarity-based Metrics} & \phantom{0} & \phantom{0} & \phantom{0} & \phantom{0} \phantom{0} & \phantom{0} & \phantom{0} & \phantom{0} & \phantom{0} & \phantom{0} \\
BLEU-1 & $0.002$ & $0.000$ & $-0.137$ & $-0.114$ & $0.104$ & $0.076$ & $0.010$ & $0.008$ & $-0.005$ & $-0.007$ \\
BLEU-2 & $-0.075$ & $-0.055$ & $0.155$ & $0.120$ & $-0.231$ & $-0.176$ & $-0.097$ & $-0.071$ & $-0.062$ & $-0.046$ \\
BLEU-3 & $-0.091$ & $-0.067$ & $0.160$ & $0.123$ & $-0.251$ & $-0.191$ & $-0.114$ & $-0.085$ & $-0.074$ & $-0.055$ \\
BLEU-4 & $-0.091$ & $-0.067$ & $0.160$ & $0.123$ & $-0.251$ & $-0.191$ & $-0.114$ & $-0.085$ & $-0.074$ & $-0.055$ \\
ROUGE-1 & $0.200$ & $0.150$ & $-0.167$ & $-0.130$ & $0.344$ & $0.251$ & $0.193$ & $0.149$ & $0.143$ & $0.105$ \\
ROUGE-2 & $0.200$ & $0.164$ & $-0.212$ & $-0.180$ & $0.359$ & $0.283$ & $0.221$ & $0.186$ & $0.142$ & $0.113$ \\
ROUGE-L & $0.202$ & $0.151$ & $-0.163$ & $-0.127$ & $0.352$ & $0.257$ & $0.199$ & $0.154$ & $0.148$ & $0.109$ \\
METEOR & $0.172$ & $0.130$ & $-0.205$ & $-0.160$ & $0.353$ & $0.260$ & $0.166$ & $0.130$ & $0.121$ & $0.090$ \\
BERTScore & $0.223$ & $0.165$ & $-0.075$ & $-0.058$ & $0.348$ & $0.256$ & $0.250$ & $0.190$ & $0.186$ & $0.139$ \\
\midrule
\textbf{Unified Evaluators} & \phantom{0} & \phantom{0} & \phantom{0} & \phantom{0} & \phantom{0} & \phantom{0} & \phantom{0} & \phantom{0} & \phantom{0} \\ 
BARTScore & $0.276$ & $0.206$ & $-0.197$ & $-0.153$ & $0.389$ & $0.283$ & $0.256$ & $0.196$ & $0.181$ & $0.133$ \\
UniEval & $0.197$ & $0.143$ & $0.034$ & $0.024$ & $0.176$ & $0.127$ & $0.213$ & $0.159$ & $0.155$ & $0.113$ \\
\midrule
\textbf{Independent Evaluators} & \phantom{0} & \phantom{0} & \phantom{0} & \phantom{0} & \phantom{0} & \phantom{0} & \phantom{0} & \phantom{0} & \phantom{0} \\
PDScore & $0.019$ & $0.013$ & \textemdash & \textemdash& $-0.106$ & $-0.074$ & $-0.005$ & $-0.004$ & $0.033$ & $0.027$ \\
Pro-Con (PC) & \textemdash & \textemdash & $0.015$ & $0.012$ & $0.051$ & $0.037$ & \textemdash & \textemdash & $0.064$ & $0.048$ \\
Argument Quality (AQ) & \textemdash & \textemdash & \textemdash & \textemdash & $0.136$ & $0.098$ & $0.084$ & $0.063$ & $0.050$ & $0.035$ \\
Toxicity Score & \textemdash & \textemdash & $0.219$ & $0.283$ & \textemdash & \textemdash & \textemdash & \textemdash & $0.087$ & $0.068$ \\
\midrule
\textbf{LLM-based Evaluators} & \phantom{0} & \phantom{0} & \phantom{0} & \phantom{0} & \phantom{0} & \phantom{0} & \phantom{0} & \phantom{0} & \phantom{0} \\
Llama3 (zero-shot) & $0.330$ & $0.286$ & $0.199$ & $0.185$ & $0.328$ & $0.279$ & $0.163$ & $0.150$ & $0.255$ & $0.225$ \\
Llama3 (G-EVAL) & $0.318$ & $0.271$ & $0.136$ & $0.126$ & $0.318$ & $0.269$ & $0.263$ & $0.239$ & $0.259$ & $0.226$ \\
Llama3 (Auto-CSEval) & $0.340$ & $0.293$ & $0.171$ & $0.159$ & $0.326$ & $0.278$ & $0.278$ & $0.252$ & $0.279$ & $0.246$ \\
Mistral (zero-shot) & $0.322$ & $0.284$ & $0.278$ & $0.254$ & $0.339$ & $0.284$ & $0.302$ & $0.277$ & $0.310$ & $0.275$ \\
Mistral (G-EVAL) & $0.360$ & $0.319$ & $0.315$ & $0.289$ & $0.366$ & $0.313$ & $0.350$ & $0.319$ & $0.348$ & $0.310$ \\
Mistral (Auto-CSEval) & $0.372$ & $0.332$ & $0.291$ & $0.268$ & $0.349$ & $0.293$ & $0.363$ & $0.333$ & $0.344$ & $0.306$ \\
GPT-4 (zero-shot) & $0.532$ & $0.476$ & $0.349$ & $0.328$ & $0.435$ & $0.377$ & $0.414$ & $0.383$ & $0.433$ & $0.391$ \\
GPT-4 (G-EVAL) & $0.626$ & $0.572$ & $0.379$ & $0.351$ & $0.582$ & $0.488$ & $0.524$ & $0.464$ & $0.527$ & $0.469$ \\
GPT-4 (Auto-CSEval) & \underline{\textbf{$0.687$}} & \underline{\textbf{$0.637$}} & \underline{\textbf{$0.447$}} & \underline{\textbf{$0.425$}} & \underline{\textbf{$0.600$}} & \underline{\textbf{$0.506$}} & \underline{\textbf{$0.567$}} & \underline{\textbf{$0.504$}} & \underline{\textbf{$0.575$}} & \underline{\textbf{$0.518$}} \\
\midrule
$\Delta_{\mathrm{GPT-4 (Auto-CSEval) (Ours)} - \mathrm{Best Method}}$ & \textcolor{ForestGreen}{\hfill$\uparrow0.062$} & \textcolor{ForestGreen}{\hfill$\uparrow0.065$} & \textcolor{ForestGreen}{\hfill$\uparrow0.068$} & \textcolor{ForestGreen}{\hfill$\uparrow0.075$} &  \textcolor{ForestGreen}{\hfill$\uparrow0.015$} & \textcolor{ForestGreen}{$\uparrow0.019$} & \textcolor{ForestGreen}{\hfill$\uparrow0.018$} & \textcolor{ForestGreen}{\hfill$\uparrow0.040$} & \textcolor{ForestGreen}{\hfill$\uparrow0.048$} & \textcolor{ForestGreen} {\hfill$\uparrow0.049$} \\
\bottomrule
\end{tabular}%
}
\end{center}
\vspace{-3mm}
\caption{Sample-level Spearman ($\rho$) and Kendall-Tau ($\tau$) correlations of different metrics on \texttt{CSEval} benchmark. For any given evaluation aspect (Relevance, for example), the correlation scores are computed against their respective human ratings. LLM-based evaluators are reported in three settings: zero-shot, chain-of-thoughts (G-EVAL), and auto-calibrated chain-of-thoughts (Auto-CSEval). Automatic and human-evaluation scores were normalised to a scale of (0,1) before computing correlation.}
\label{tab:results_table}
\vspace{-5mm}
\end{table*}

\section{Experimental Setup}

\subsection{Evaluation Metrics}
\label{sec:evaluation_metrics}

We benchmark \texttt{CSEval} on traditional similarity-based metrics, as well as multiple popular single-dimensional (independent), unified, and LLM-based evaluators.  

(i) {\bf Similarity-based metrics} measure the degree of lexical or semantic overlap between the predicted and reference counterspeech. Specifically, we report \textbf{{BLEU}} \cite{papineni2002bleu}, \textbf{{ROUGE}} \cite{rouge-score-lin-2004}, \textbf{{METEOR}} \cite{meteor-score-banerjee-lavie-2005}, and \textbf{{BERTScore}} \cite{bert-score-zhang2020}, the most widely reported metrics in the counterspeech literature.

(ii) {\bf Single-dimensional evaluators} include some popular models that score uni-dimensional quality attributes of a counterspeech, such as degree of toxicity, quality of arguments, etc. We report \textbf{{Pro-Con}} \cite{bar-haim-etal-2021-project-debator}, \textbf{{Argument-Quality}} \cite{bar-haim-etal-2021-project-debator}, and \textbf{{Toxicity}} \cite{Detoxify} pretrained classifiers evaluating the stance, quality of arguments, and degree of toxicity in a counterspeech respectively. Further, we include \textbf{{PD-Score}} \cite{halim2023wokegpt}, which evaluates the effectiveness of counterspeech. 

(iii) {\bf Unified evaluators} are neural networks trained to predict aspect-level scores for a given text. We report {\textbf{BARTScore}}, which is a unified evaluator that uses the average likelihood of the model output as the metric. We also report {\textbf{UniEval}} \cite{zhong-etal-2022-UniEVAL}, which uses a pre-trained T5 model to encode the evaluation task. It encodes source and target texts as question-answer pairs and then computes the QA score as the evaluation score. 

(iv) {\bf LLM-based evaluators} are methods that leverage LLMs to assess the quality of the generated text using natural language instructions or automated prompting techniques. We include zero-shot baselines of \textbf{Llama} \cite{touvron2023_llama} \textbf{Mistral} \cite{jiang2023_mistral}, and \textbf{GPT-4} \cite{ouyang2022-instuctGPT}. Furthermore, we  report
\textbf{G-EVAL} \cite{liu2023-GEVAL}, an auto-CoT prompting framework, and the current state-of-the-art in NLG evaluation. 
\section{Results}

We adopt the same approach suggested by \citet{zhong-etal-2022-UniEVAL} and \citet{liu2023-GEVAL} to evaluate different \texttt{CS} generation metrics using \texttt{CS}-level Spearman ($\rho$) and Kendall-Tau ($\tau$) correlation. We report the correlation scores between automated metrics and human judgments in Table \ref{tab:results_table}. The first part of Table \ref{tab:results_table} shows metrics that compare lexical or semantic similarity between the model output and reference \texttt{CS}. We find that \textbf{overlap-based metrics perform poorly on almost all dimensions}, with metrics such as BLEU, ROUGE, and METEOR even displaying negative correlations in some cases. This observation is consistent with similar studies in summarisation and dialogue generation, where overlap-based metrics are shown to be poor multidimensional evaluators \cite{fabbri2021summeval, zhong-etal-2022-UniEVAL, liu2023-GEVAL}. The second part shows results of metrics that use neural networks to learn from human ratings of text quality. In particular, BERTScore and BARTScore show a higher correlation with human judgement than all other similarity-based metrics. This shows that they are more reliable for \texttt{CS} evaluation.

The third part of Table \ref{tab:results_table} shows the results of some popular models used for \texttt{CS} evaluation. These models independently evaluate specific quality aspects of a \texttt{CS}. We observe that \textbf{most independent evaluators display poor correlation with human judgements for the aspect that they evaluate}. While the Toxicity score \cite{Detoxify} shows moderate correlation, it still lags behind LLM-based evaluators in assessing \textit{aggressiveness} of a \texttt{CS}. 

Lastly, we observe that \textbf{LLM-based evaluators achieve the highest correlations with human judgements across all dimensions}, underlying their reliability as reference-free counterspeech evaluators. Our proposed {Auto-CSEval} method with GPT-4 as the base model consistently outperforms all the other LLM baselines, especially for \textit{relevance}, \textit{coherence}, and \textit{suitableness}. Furthermore, we observe that for each of the three LLMs, prompting methods with CoT (G-EVAL and Auto-CSEval) perform much better than their respective zero-shot counterparts. This validates our hypothesis that CoT in the form of evaluation steps helps LLMs achieve improved alignment, bringing them closer towards human reasoning for counterspeech evaluation.
\section{Analysis}
\label{sec:analysis}

\paragraph{Effect of Calibration.} In Table \ref{tab:results_table}, we compare the performance of Auto-CSEval with and without auto-calibrated CoT on the \texttt{CSEval} benchmark.  shows that Auto-CSEval (GPT-4) outperforms G-EVAL (GPT-4) on both Spearman and Kendall-Tau correlations. We observe a similar trend for Mistral and Llama models. This suggests that auto-calibration helps align the CoT towards human judgement. 

\paragraph{Is there a case for a single, unified score to assess counterspeech quality?} Our proposed framework is designed to independently evaluate a \texttt{CS} across different quality dimensions. This is in line with the broader counterspeech literature, which supports the notion that there is no single, absolute metric or aspect that completely represents a counterspeech's quality or effectiveness \cite{benesch2014countering, chung-et-al-2023-understanding-counterspeech}. However, research following \texttt{CS} generation often requires to relatively compare different NLG methods. To aid this, we experiment with the efficacy of having a unified score for \texttt{CS} evaluation. Specifically, for a given model output, we compute the unified score as a mean of individual scores -- relevance, coherence, aggressiveness, and suitableness. 

\small
\begin{align}
    \text{Mean} &= \frac{ \text{Relevance} + (6 - \text{Aggressiveness}) + \text{Coherence} }{4} \notag \\
    &\quad + \frac{ \left( \frac{\text{Suitableness} - 1}{2} \times 4 + 1 \right) }{4}
\end{align}
\normalsize

As shown in the equation above, Aggressiveness score is normalised as \(6 - \text{Aggressiveness}\) since it is a "lower the better" score. Similarly, Suitableness is transformed from the original scale of 1-3 to 1-5. This is the same logic used to compute scores in Figure \ref{fig:introduction_figure}. 

For this experiment, we construct a meta-evaluation by randomly sampling model outputs from \texttt{CSEval}. For a given \texttt{HS}, we ask human evaluators to rank model outputs from best to worst according to their preference\footnote{
\textcolor{black}{We conduct the meta-evaluation on 372 randomly selected data points from the \texttt{CSEval} test set, representing approximately 15\% of the total set. Each data point was independently ranked by three human annotators, and the final rankings were determined by averaging the individual rankings from the three annotators}.}. Table \ref{tab:system_level_performance_table} shows the system-level ranking performance of Llama, Mistral, and GPT-4, computed using the Normalized Discounted Cumulative Gain (NDCG) score. We observe that all LLM-based methods show excellent ranking performance, displaying alignment with human preferences. These results support the use of a unified score, indicating that it can be effective in comparing the relative quality of \texttt{CS} generation models.  
\begin{table}[t!]
\begin{center}
\small
\resizebox{\columnwidth}{!}{%
\begin{tabular}{l|c}
\toprule
\multicolumn{1}{l|}{\textbf{Model}} & {\textbf{System-level preference ranking}} \\
\midrule
LLama-3 (GEval) & $0.874$ \\
Mistral (GEval) & $0.862$ \\
GPT-4  (GEval) & $0.894$ \\
LLama-3 (Auto-CSEval) & $0.890$ \\
Mistral (Auto-CSEval) & $0.891$ \\
GPT-4  (Auto-CSEval) & $0.898$ \\
\bottomrule
\end{tabular}%
}
\end{center}
\caption{\textcolor{black}{The performance of various methods in system-level preference ranking evaluated using the \textbf{NDCG score}, which ranges from 0 to 1. Notably, all LLM-based evaluators demonstrate exceptional effectiveness in ranking model outputs.}}
\label{tab:system_level_performance_table}
\vspace{-5mm}
\end{table}


\section{Conclusion}
This paper presents \texttt{CSEval}, a novel multidimensional framework for evaluating the quality of automated counterspeech against online hate speech. Our work addresses a critical gap in the current research landscape of automated counterspeech generation by providing a comprehensive and automated approach for assessing counterspeech quality along four key dimensions: context relevance, aggressiveness, argument coherence, and suitableness. We observe that traditional similarity-based evaluation metrics, while prevalent, often fail to capture the nuanced complexity of effective counterspeech. We further introduce \texttt{Auto-CSEval}, a prompt-based evaluation method with an auto-calibrated chain-of-thoughts mechanism, which leverages LLMs to offer a more refined and human-aligned evaluation. Our experiments with multiple automated metrics on the \texttt{CSEval} dataset illustrate that Auto-CSEval displays a significant improvement in correlation with human judgment, particularly when juxtaposed against existing evaluation methods. In conclusion, while our framework marks a significant advance in the automated evaluation of counterspeech, it also underscores the complexity and multi-faceted nature of this domain. 
\section{Limitations}

Our study focuses on four dimensions of text quality that are specific to counterspeech generation: relevance, coherence, aggressiveness, and appropriateness. However, there are other quality aspects, such as fluency, naturalness, opposition (stance), and specificity \cite{chung-et-al-2023-understanding-counterspeech, jones2024-multiaspectframeworkcounternarrative}. In future work, we plan to extend CSEval to incorporate these quality dimensions, since it can add value to the notion of "unified score" as discussed in section \ref{sec:analysis}. Recent studies have shown that using LLMs as reference-free evaluators (LLMs-as-a-judge) may lead to unseen problems like bias, prompt sensitivity, leniency, and questionable reasoning \cite{thakur2025judgingjudgesevaluatingalignment, li2025generationjudgmentopportunitieschallenges}. Furthermore, LLM evaluations may also be affected by decoding strategies (temperature or top-k sampling). Our study does not include any analysis on such weakness of LLMs-as-a-judge. Lastly, we would also like to explore the trade-offs between different text quality dimensions and how they affect the perception and reception of counterspeeches. We leave this to future work. 
\section{Ethics Statement}
We acknowledge that we are dealing with a sensitive topic of research as it deals with online hate speech. We understand that we must adhere to strict ethical considerations while dealing with hatespeech-related data. First, our dataset is based on a publicly available, open-source dataset in the counterspeech domain. As we were largely dealing with online hate speech in the form of social media posts, we ensure that they were fully anonymised and untraceable to the source user. During the data annotation process, we ensured that each of the annotators was fully aware and had context about the nature and degree of offensive statements that they were responsible to annotate. 
\section{Acknowledgments}
We extend our gratitude to our students Shaily Desai, Osho Anand, and Apporv Jain for their active participation during data curation and the central HPC facility (Padum) at IIT Delhi for computing. We also sincerely thank Logically and Anusandhan National Research Foundation
(CRG/2023/001351) for financial support. Tanmoy acknowledges the support of Rajiv Khemani Young Faculty Chair Professorship in Artificial Intelligence. 

\bibliography{main}

\appendix
\begin{figure*}[t]
\includegraphics[width=\textwidth]{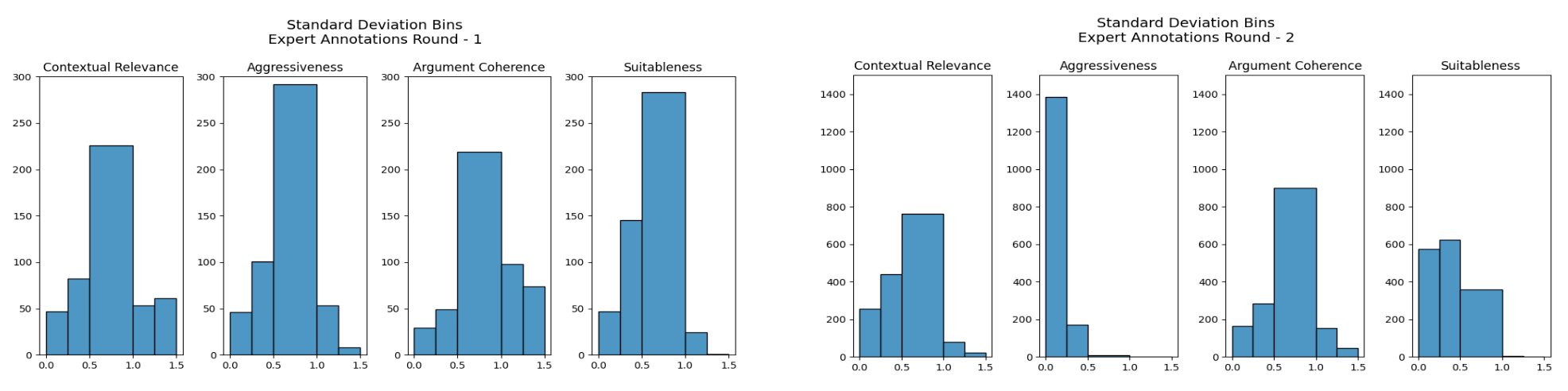}
\caption{Histograms of standard deviations of inter-annotator scores between first-round expert annotations and second-round expert annotations.}
\label{fig:histogram}
\end{figure*}
\begin{table*}[!ht]
\begin{center}
\small 

\label{tab:overall_counter_speech_examples}

\begin{subtable}{\textwidth}
\label{subtab:part_a}
\begin{tabularx}{\textwidth}{p{12cm}cX} 
\toprule
\textbf{Generated Counterspeech} & \textbf{Model} & \textbf{Expert Scores (avg)} \\
\midrule
\addlinespace
It's vital for us to approach each other with understanding and empathy. While economic concerns can be challenging for everyone, it's important to remember that people of all backgrounds contribute to our society in many ways. Rather than focusing on divisions, let's strive to recognize our shared human experiences and the value of diverse contributions. By fostering a sense of community and understanding, we can create a more inclusive and supportive environment for everyone, regardless of their background. & M1 & Rel: 1.4 Agg: 1.1 Coh: 4.8 Sui: 1.5 \\
\midrule
\addlinespace
Your statement is blatantly racist and ignorant. People of every race can face legal or illegal immigration statuses, and work ethic is not defined by skin color. Spouting such nonsense only reveals a deep-seated prejudice and lack of understanding about the realities of our diverse society. & M2 & Rel: 3.4 Agg: 4.5 Coh: 2.1 Sui: 1.0 \\
\midrule
\addlinespace
Is it really fair to claim that an entire racial group is here illegally and unable to work? & M3 & Rel: 2.1 Agg: 1.5 Coh: 1.0 Sui: 1.0 \\
\midrule
\addlinespace
Being in a country illegally is not exclusive to any race, including Black people. Legal status depends on individual circumstances, not race. Secondly, statistical evidence shows that Black individuals are actively contributing to the workforce in various sectors. The claim that they 'cannot work' is unsupported by data. Additionally, the need for financial assistance is a situation that can affect anyone, regardless of their racial background. Misrepresenting these facts not only perpetuates harmful stereotypes but also undermines the reality of our diverse and collaborative society. & M4 & Rel: 4.6 Agg: 1.5 Coh: 4.8 Sui: 3.0 \\
\bottomrule
\end{tabularx}
\caption{}
\end{subtable}

\vspace{1em} 

\begin{subtable}{\textwidth}
\label{subtab:part_b}
\begin{tabularx}{\textwidth}{p{12cm}cX} 
\toprule
\textbf{Reference Counterspeech} & \textbf{Model} & \textbf{Expert Scores (avg)} \\
\midrule
\addlinespace
Firstly, the legality of a person's presence in a country is governed by immigration laws, not their race. People of all races can be found in both legal and illegal statuses. Second, employment statistics clearly show that Black individuals are an active and integral part of the workforce in various sectors. The idea that they 'cannot work' is simply untrue. Lastly, the need for financial support is a circumstance that can affect anyone, regardless of race. It's important we base our discussions on facts and evidence, not stereotypes. Spreading inaccurate information only serves to deepen divisions and misunderstandings in our society. & - & Rel: 4.1 Agg: 2.5 Coh: 4.0 Sui: 3.0 \\
\midrule
Your views on Black people are based on myths. Legal status isn't defined by race. Many Black individuals work hard and contribute to our economy. It's unfair and incorrect to stereotype them as you did. & - & Rel: 5.0 Agg: 1.0 Coh: 5.0 Sui: 3.0 \\
\bottomrule
\end{tabularx}
\caption{}
\end{subtable}
\caption{Overview of data from the CSEval dataset. We provide expert human annotations spanning four quality aspects of counterspeech, rated for outputs from different counterspeech models - contextual-relavance (Rel), aggressiveness (Agg), argument-coherence (Coh), and suitableness (Sui).}
\end{center}
\end{table*}

\begin{algorithm*}[!ht]
\caption{Calibrating CoT Evaluation Steps Using Auto-CSEval}
\label{tab:pseudo_code}
\begin{algorithmic}

\Require LLM model $\theta$, human judgments dataset $H$, correlation metric $f(\cdot)$, Monte Carlo trials $N$, set of few-shot example sizes $L = \{l_1, l_2, \ldots, l_m\}$, evaluation aspect $a$, target CoT candidate pool size $k$
\\
\Statex
\Comment{Iterate over few-shot example sizes}
\For{$l_i \in L$}
    \Comment{Perform $N$ Monte Carlo trials}
    \For{$j = 1$ to $N$}
        \State Sample few-shot examples $H_s = \bigcup (h_i, s_i, a)$ from $H$
        \State Generate CoT candidate using $\theta$ with temperature sampling
        \State Add CoT candidate $C_i$ to global set $C$
    \EndFor
\EndFor

\Statex
\Comment{Select top-$k$ CoT candidates based on correlation}
\State $C \leftarrow \text{Top-}k\{c_i \in C \mid f(c_i, H)\}$

\Statex
\Comment{Refine misaligned or low-scoring candidates}
\For{$c_i \in C$}
    \State Collect misaligned examples $H_{M_i}$ for $c_i$
    \For{$j = 1$ to $N$}
        \State Sample examples $H_{M_s} = \bigcup (h_{M_i}, s_{M_i}, a)$ from $H_{M_i}$
        \State Refine candidate CoT with $\theta$ and add to $C$
    \EndFor
\EndFor

\Statex
\Comment{Return the best-calibrated CoT}
\State \Return Calibrated CoT $C_f \leftarrow \arg\max_{c_i \in C} f(c_i, H)$

\end{algorithmic}
\end{algorithm*}

\begin{figure*}[t]
\centering
\includegraphics[width=0.9\textwidth]{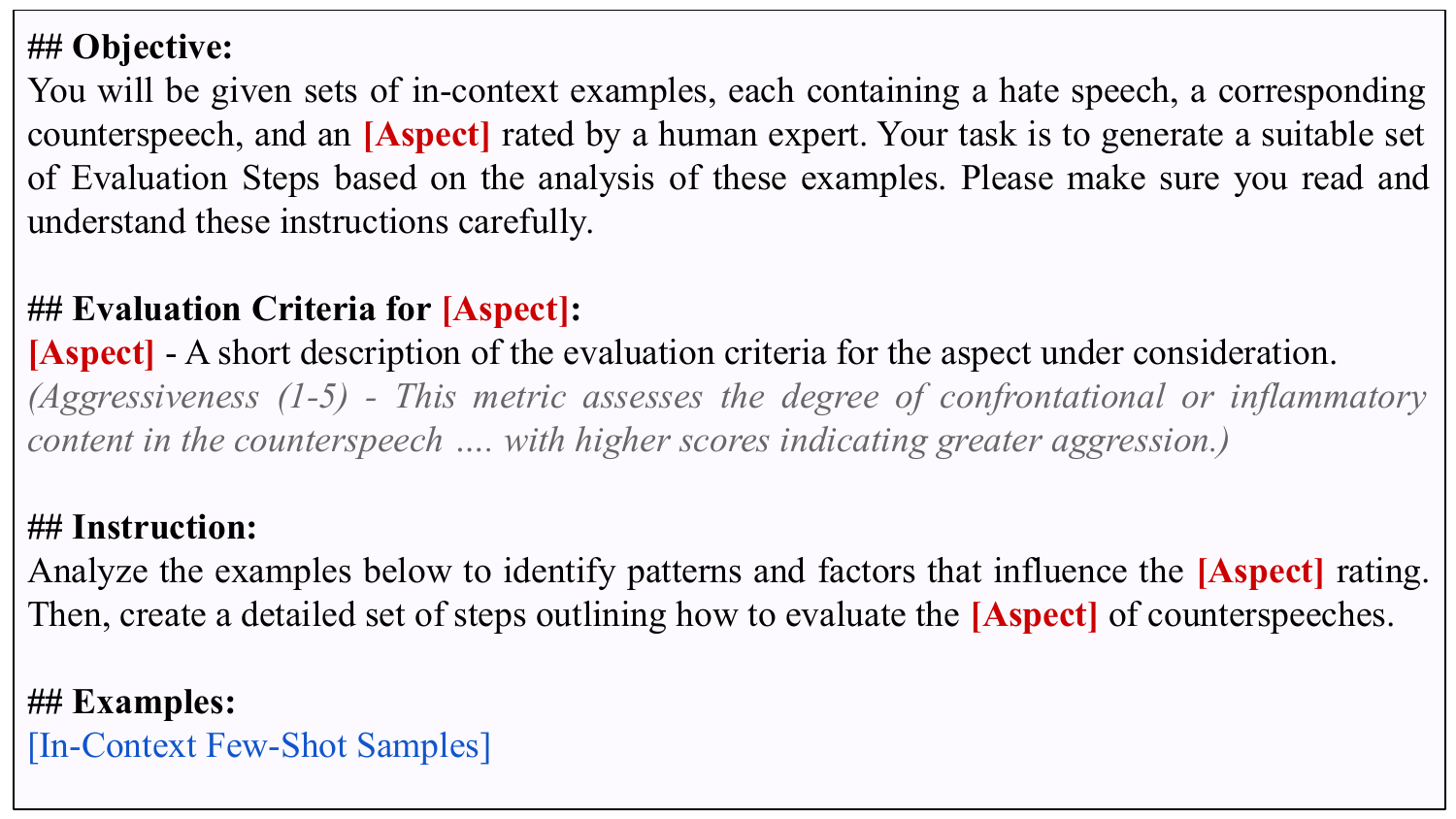}
\caption{Prompt template: Candidate CoT drafting.}
\label{fig:prompt_template_cot_drafting}
\end{figure*}
\begin{figure*}[t]
\includegraphics[width=\textwidth]{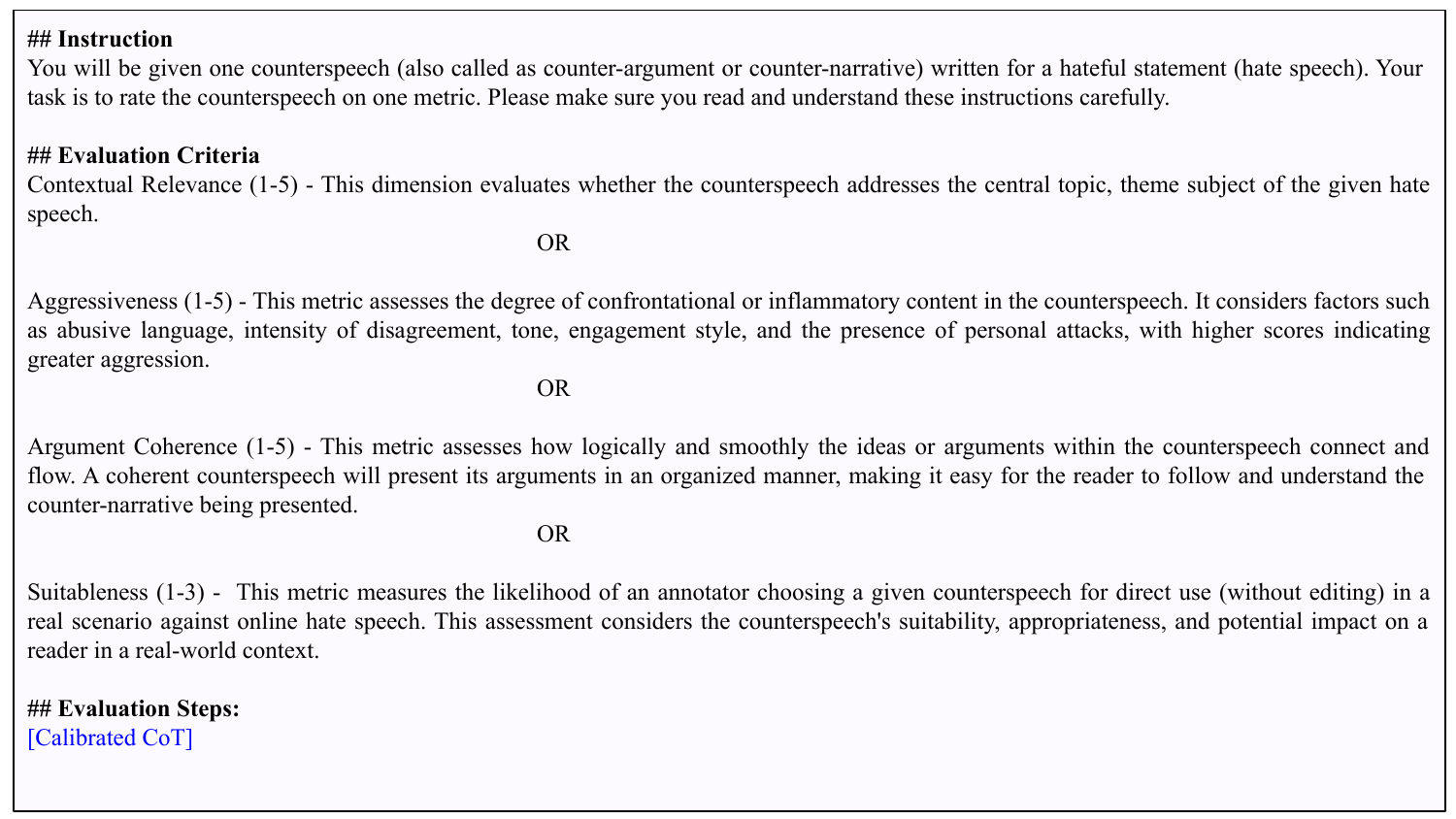}
\caption{Prompt template: Scoring a counterspeech.}
\label{fig:prompt_template_scoring}
\end{figure*}
\begin{figure*}[t]
\includegraphics[width=\textwidth]{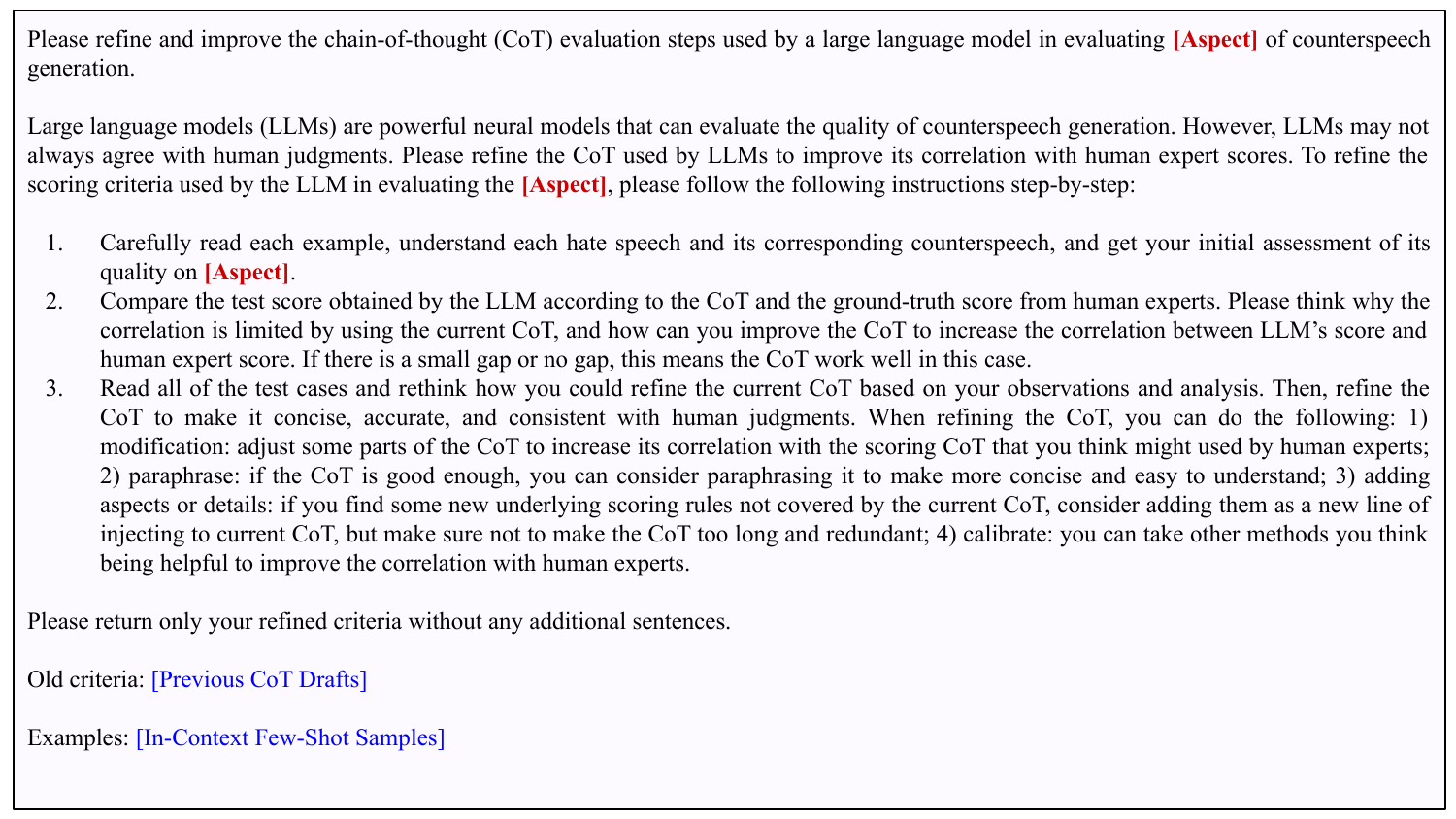}
\caption{Prompt template: Candidate CoT refinement.}
\label{fig:prompt_template_refinement}
\end{figure*}
\newpage

\section{Dataset}
\label{sec:appendix-dataset}


\subsection{Annotation Process} We recruited five expert annotators who either published papers or completed senior theses in the domain of hate speech and counterspeech. We used expert annotators for our evaluation process, due to the task's sensitivity and the quality issues of crowd-sourced annotations reported in previous work \cite{gillick-liu-2010-nonExpertAnnotationsRisky}. Before starting the annotation process, we conducted a joint discussion session among the annotators. Here, we reviewed resources such as the field manual for responding to online abuse \footnote{\href{https://onlineharassmentfieldmanual.pen.org/.}{https://onlineharassmentfieldmanual. pen.org/.}} with the goal of achieving a common understanding of the task.

We designed a simple data annotation interface that showed the annotators a hate speech statement and three model-generated counterspeeches randomly grouped together. Each group of model outputs included the reference counterspeech to provide a common point of reference across groups. Annotators were then asked to rate the counterspeech on a Likert scale for all four dimensions as described in the main text.

We conducted three rounds of annotation to ensure the high quality and reliability of the annotations. The first round involved three annotators rating a randomly sampled set of 500 examples for each of the four dimensions. At the end of the first round, annotators were asked to review odd-man-out cases, i.e., examples where their score of a dimension differed from other annotators by more than 2 points. For cases where such a pattern did not exist, all annotators examined the annotation. When re-rating examples, annotators could see the scores given by other expert annotators in the first round of annotations. Although this setting could bias the re-assigned scores towards the average judgement from the first round, we urged experts to be critical and discuss disputed examples when needed. We found that these discussions helped improve agreement across all dimensions in the second round, where the annotators rated another randomly sampled set of 1000 examples. Finally, after achieving good agreement scores in the second round, the third round was done independently, where each annotator rated a separate set of examples.

\subsubsection{Supervised Methods} 
\label{appendix:model_description}
We train three state-of-the-art models for counterspeech generation using the IntentConan train set. For each model, we detail the specific hyperparameters used during the fine-tuning process below. 

\textbf{M1 - Generate Prune Select (GPS)} \cite{zhu-bhat-2021-generate}: This model employs a three-step process of generating, pruning, and selecting counterspeeches using an autoencoder, a grammatical filter, and a vector similarity measure. \textcolor{black}{We fine-tune the GPS model on the \texttt{IntentCONAN} train set with the following hyperparameters: \textit{learning rate} = $\texttt{1e-3}$, \textit{batch size} = $8$, \textit{number of epochs} = $500$, \textit{mixed precision} = \texttt{False}}.

\textbf{M2 - DialoGPT} \cite{zhang2020dialogpt}: DialoGPT is a large-scale pre-trained language model that excels in generating context-aware responses, surpassing models like GPT-2. \textcolor{black}{Specifically, we use the \texttt{microsoft/DialoGPT-small} version of DialoGPT. We fine-tune DialoGPT on the \texttt{IntentCONAN} train set with the following hyperparameters: \textit{learning rate} = $\texttt{5e-5}$, \textit{batch size} = $4$, \textit{number of epochs} = $3$, \textit{mixed precision} = \texttt{False}}.

\textbf{M3 - QUARC} \cite{gupta-etal-2023-counterspeeches}: QUARC represents the state-of-the-art in generating counterspeeches tailored to specific intents. \textcolor{black}{We fine-tune the QUARC model on the \texttt{IntentCONAN} train set with the following hyperparameters: \textit{learning rate} = $\texttt{5e-5}$, \textit{batch size} = $4$, \textit{number of epochs} = $3$, \textit{mixed precision} = \texttt{False}}.

\subsubsection{Prompt-Based Methods} These methods use natural language prompts to generate counterspeeches from pre-trained language models without any fine-tuning. We include the following two models in this category:

M4 - \textbf{ChatGPT (GPT3.5-turbo)} \cite{gupta-etal-2023-counterspeeches} uses a single prompt to generate counterspeeches for different intents from a large-scale pre-trained language model. \textcolor{black}{Specifically, we use the \texttt{gpt-3.5-turbo} model version}. We include both zero- and few-shot strategies of prompting. 

M5 - \textbf{GPT-4} \cite{gupta-etal-2023-counterspeeches} uses a few examples of counterspeeches for each intent to guide the generation of new counterspeeches from a large-scale pre-trained language model. \textcolor{black}{Specifically, we use the \texttt{gpt-4} model version}. We include both zero- and few-shot strategies of prompting.

To provide relevant ICL examples for few-shot prompting of GPT-3.5 and GPT-4, we followed these steps:
\begin{enumerate}[nosep]
    \item 
    \textcolor{black}{We identify semantically similar instances to the input hatespeech from the IntentCONAN dataset. We used these similar instances as in-context learning (ICL) examples in the prompt.}
    
    \item 
    \textcolor{black}{    We use the  \texttt{all-mpnet-base-v1} model from the sentence-transformers library \cite{sentence-transformers} to retrieve the semantically similar examples. This model calculates similarity scores between the input hatespeech and the instances in the dataset.}
    
    \item 
    \textcolor{black}{ We select the top three instances with the highest similarity scores. These three most relevant examples were then included in the prompt for few-shot learning.}

\end{enumerate}

\section{Prompting Templates}
\label{appendix:prompt_templates}
In this section, we list prompt templates applied throughout this study, including induction templates for CoT criteria drafting, evaluation templates that utilize the generated CoT for scoring, and templates for self-refinement of CoT.

\subsection{CoT Drafting Templates}
Prompt templates for CoT prompt drafting are listed in Figure \ref{fig:prompt_template_cot_drafting}. The [Aspect] denotes placeholders for aspects to evaluate (e.g., coherence, relevance, etc.), and sampled few-shot in-context exemplars are placed at [In-Context Few-Shot Samples], including samples and their expert scores.

\subsection{Evaluation Templates}
Prompt templates for evaluation are listed in Figure \ref{fig:prompt_template_scoring}. The [Aspect] denotes placeholders for aspects to evaluate (e.g., coherence, aggressiveness, etc.). Evaluation samples and calibrated CoT prompts for each aspect are filled into corresponding placeholders during evaluation.

\subsection{CoT Refinement Templates}
An example prompt template for CoT prompt refinement can be found in Figure \ref{fig:prompt_template_refinement}. As illustrated in the figure, we first fill in the aspect and tasks to the instructions, then prompt the LLM with the previous CoT, few-shot in-context samples of misaligned evaluations, together with suggested means of modifications to obtain a modified version of scoring criteria for this task.


\section{Evaluation Strategy}
\label{sec:evaluation_strategy}

Assume a set of conditioned text (reference counterspeech) $\{c_1, c_2, ..., c_n\}$ and $M$ NLG models. The generated text of $m$-th model for the $i$-th condition text is denoted as $g_{i,m}$. Sample-level evaluation strategy computes the correlation scores as follows: \begin{equation}
\begin{aligned}
\text{Corr}_{\text{sample}} = \frac{1}{n} \sum_{i=1}^{n} (\rho([f_{\text{auto}}(g_{i,1}), ..., f_{\text{auto}}(g_{i,M})], \\ [f_{\text{human}}(g_{i,1}), ..., f_{\text{human}}(g_{i,M})]))
\end{aligned}
\end{equation}  
where $\rho$ denotes Spearman correlation, and $f_{\text{auto}}$ and $f_{\text{human}}$ indicate the automatic and human evaluation functions, respectively.

\subsection{Select of Evaluation Dimensions}
\label{sec:appendix_eval_dims}
\textcolor{black}{Our rationale for selecting the evaluation dimensions — relevance, aggressiveness, coherence, and suitability was based on their frequent reporting in existing literature. In Table \ref{tab:evaluation_dimensions}, we highlight each evaluation dimension and the studies that report them in their human evaluations.}

\textcolor{black}{Suitableness is scored a Likert scale of 1-3 because we found that it more appropriate given it's definition - its also similar to
-or-Not used by \cite{tekiroglu-etal-2022-using}. All the other dimensions are scored on a Likert scale of 1-5.}

\textcolor{black}{Furthermore, our decision to adopt a Likert (1-5) scoring scale was informed by past research in curating NLG evaluation benchmarks. This includes well-established benchmarks such as SummEval \cite{fabbri2021summeval}, Topical Chat \cite{gopalakrishnan2023topicalchat}, and FED \cite{mehri-eskenazi-2020-unsupervised}. In fact, in contemporary literature, discrete, ordinal Likert scales are the most commonly used method for human evaluation of NLG \cite{wang2023chatgpt_a_good_NLG_evaluator, liu2023-GEVAL, fu2023GPTScore}
}

\begin{table*}[t!]
\centering
\resizebox{\textwidth}{!}{%
\begin{tabular}{|l|p{8cm}|p{4cm}|}
\midrule
\textbf{Evaluation Dimension} & \textbf{Brief Description} & \textbf{References} \\ 
\midrule
\textbf{Contextual Relevance} & \textcolor{black}{Assesses how well the counterspeech relates to the original hate speech. Ideally, a counterspeech must be contextually relevant or "on-topic" with the hatespeech.} & \cite{chung-etal-2021-towards} \newline \cite{halim2023wokegpt} \newline \cite{gupta-etal-2023-counterspeeches} \\ \midrule
\textbf{Argument Coherence} & \textcolor{black}{Evaluates the logical flow and consistency of the counterspeech. An idean counterspeech must present a clear and convincing argument.} & \cite{tekiroglu-etal-2022-using} \newline \cite{hengle2024intentconditioned} \\ 
\midrule
\textbf{Aggressiveness} & \textcolor{black}{Measures the degree of hostility or confrontational tone in the counterspeech. Ideally, a counterspeech must maintain a respectful and non-aggressive discourse.} & \cite{ashida-komachi-2022-towards}  \newline \cite{gupta-etal-2023-counterspeeches} \newline \cite{hengle2024intentconditioned}  \\ 
\midrule
\textbf{Suitability} & \textcolor{black}{Evaluates how appropriate or effective a counterspeech is in aligning with social norms and the specific context.} & \cite{chung-etal-2021-towards} \newline \cite{bonaldi-etal-2022-human} \newline \cite{tekiroglu-etal-2022-using} \\ 
\end{tabular}%
}
\caption{\textcolor{black}{An overview of selected evaluation dimensions in CSEval, along with their mentions in human evaluation studies conducted by contemporary literature.}}
\label{tab:evaluation_dimensions}
\end{table*}

\section{Statistical Tests}

\begin{table*}[ht]
    \centering
    \label{tab:sr}
    \resizebox{\textwidth}{!}{ 
    \begin{tabular}{lcccccccc}
        \toprule
        \textbf{Model} & \multicolumn{2}{c}{\textbf{Relevance Score}} & \multicolumn{2}{c}{\textbf{Aggressiveness Score}} & \multicolumn{2}{c}{\textbf{Coherence Score}} & \multicolumn{2}{c}{\textbf{Suitableness Score}} \\
        \cmidrule(lr){2-3} \cmidrule(lr){4-5} \cmidrule(lr){6-7} \cmidrule(lr){8-9}
        & Mean & Std & Mean & Std & Mean & Std & Mean & Std \\
        \midrule
        BLEU\_1 & -0.007 & 0.022 & -0.147 & 0.043 & 0.115 & 0.040 & 0.005 & 0.025 \\
        ROUGE\_l & 0.213 & 0.037 & -0.167 & 0.020 & 0.354 & 0.027 & 0.200 & 0.027 \\
        METEOR\_score & 0.178 & 0.032 & -0.207 & 0.020 & 0.355 & 0.025 & 0.169 & 0.023 \\
        BERT\_score & 0.227 & 0.027 & -0.077 & 0.027 & 0.342 & 0.020 & 0.249 & 0.014 \\
        Toxicity & -0.019 & 0.030 & 0.272 & 0.018 & -0.184 & 0.037 & -0.107 & 0.028 \\
        BART\_score & 0.272 & 0.024 & -0.198 & 0.023 & 0.394 & 0.024 & 0.249 & 0.021 \\
        GPT-4 (zeroshot) & 0.529 & 0.031 & 0.336 & 0.061 & 0.444 & 0.025 & 0.422 & 0.039 \\
        GPT-4 (GEval) & 0.629 & 0.012 & 0.392 & 0.030 & 0.578 & 0.031 & 0.522 & 0.014 \\
        GPT-4 (Auto-CSEval) & 0.687 & 0.010 & 0.461 & 0.035 & 0.597 & 0.032 & 0.566 & 0.014 \\
        \bottomrule
    \end{tabular}
    }
    \caption{Spearman ($\rho$) correlation results for selected methods based on statistical testing.}
\end{table*}

\begin{table*}[t]
    \centering
    \label{tab:sd}
    \resizebox{\textwidth}{!}{ 
    \begin{tabular}{lcccc}
        \toprule
        \textbf{Model} & \multicolumn{1}{c}{\textbf{Relevance Score}} & \multicolumn{1}{c}{\textbf{Aggressiveness Score}} & \multicolumn{1}{c}{\textbf{Coherence Score}} & \multicolumn{1}{c}{\textbf{Suitableness Score}} \\
        \cmidrule(lr){2-2} \cmidrule(lr){3-3} \cmidrule(lr){4-4} \cmidrule(lr){5-5}
        & Mean & Mean & Mean & Mean \\
        \midrule
        Auto-CSEval - Best Traditional Metric & 0.415 & 0.189 & 0.202 & 0.317 \\
        Auto-CSEval - Best Method & 0.058 & 0.069 & 0.019 & 0.044 \\
        \bottomrule
    \end{tabular}
    }
    \caption{Delta values comparing Auto-CSEval against the best traditional metric and best method across evaluation aspects.}
\end{table*}

To verify the reliability of the correlation scores reported in Table \ref{tab:results_table}, we conduct statistical tests for Spearman ($\rho$) correlation values by averaging results across multiple trials. The details of our experimental setup are as follows:

\begin{itemize}
    \item Table \ref{tab:results_table} reports sample-level Spearman ($\rho$) and Kendall-Tau ($\tau$) correlations of different metrics on the CSEval test set. These results are averaged over the entire dataset.
    \item We define a ``trial'' using two variables: sample size ($n$) and random state ($s$). In each trial, we randomly sample $n$ data points from the CSEval test set using the specified random state $s$.
    \item We vary the sample sizes linearly, starting from 100 samples and increasing by 500 samples at each step, i.e., 100, 600, 1100, and so on, up to the size of the test set (13 values in total).
    \item Following this, we compute correlation scores (same as Table \ref{tab:results_table}) on the randomly sampled subset ($n$ data points) of CSEval.
    \item We repeat this procedure for three different seed values: [1, 2, 3].
\end{itemize}

Thus, we conduct a total of 39 unique trials (13 sample sizes, $\times$ 3 random states). We then average the scores across all 39 trials to finalise our results. The spearman ($\rho$) correlation results for selected methods are presented in Table \textbf{5}.

In Table \textbf{6}, we calculated the delta values between (i) Auto-CSEval versus the best traditional metric and (ii) Auto-CSEval versus the best method. These delta values provide a statistically more reliable comparison than those reported in Table 1 of the original manuscript, as the results are averaged across 13 sample sizes and three seed values. We find that \textbf{our proposed method, Auto-CSEval, shows an average improvement of 0.190 points over the best traditional metric}. This improvement is consistent across four evaluation aspects: relevance, coherence, aggressiveness, and suitableness.

\end{document}